\documentclass{article}
\usepackage{spconf,amsmath,graphicx,hyperref}

\usepackage{booktabs}
\usepackage{adjustbox}
\usepackage{amssymb}
\usepackage{multirow} 
\usepackage{hyperref}

\title{Beyond Face Swapping: A Diffusion-Based Digital Human
Benchmark for Multimodal Deepfake Detection}
\name{Jiaxin Liu$^{\dag}$, Jia Wang$^{\dag}$, Saihui  Hou$^{\star}$, Min Ren, Huijia Wu, Long Ma, Renwang Pei, Zhaofeng He$^{\star}$
\thanks{$^{\dag}$ These authors contribute equally to this work. $^{\star}$ Corresponding author. This work was supported in part by the National Natural Science Foundation of China (Grants 62576046, 62301066, and 62406028), the Key Project of Philosophy and Social Sciences Research of the Ministry of Education of China (No. 24JZD040), the Fundamental Research Funds for the Central Universities (2023RC72), and Beijing University of Posts and Telecommunications (2025YZ010). And the work is based on the FlagEval system of BAAI.
}
}
\address{BUPT, BNU, Didi Chuxing, Beijing, China}
\begin{document}
\topmargin=0mm
%
\maketitle
\begin{abstract}
Recent advances in deepfake technologies, particularly digital human generation, pose escalating threats due to their realism and covert multimodal control. Existing detection methods struggle with these new challenges. To address this gap, we first present \textbf{DigiFakeAV}\footnote{Project:\quad\url{https://hubeiwuhanliu.github.io/DigiFakeAV.github.io/}}, a large-scale, multimodal forgery dataset featuring 60,000 videos generated by five state-of-the-art digital human models, ensuring diversity in nationality, skin tone, gender, and scenario. Existing detectors exhibit marked performance drops on DigiFakeAV, underlining its difficulty. To counter this, we propose DigiShield, a robust baseline leveraging spatiotemporal and cross-modal fusion of visual and audio features. DigiShield attains state-of-the-art results on DigiFakeAV and generalizes well to other datasets.

\vspace{-4pt}

\end{abstract}

\section{Introduction}
\label{sec:intro}
The rapid advancement of deepfake technology has posed unprecedented threats to digital security, giving rise to malicious applications such as political disinformation and sophisticated financial fraud. While existing detection benchmarks have driven progress, they remain constrained by outdated synthesis paradigms. The existing deepfake detection datasets can be roughly categorized into three generations~\cite{dolhansky2020deepfakedetectionchallengedfdc}. First-generation datasets (e.g., FF++~\cite{rössler2019faceforensicslearningdetectmanipulated}) suffer from limited scale and quality. The second generation (e.g., Celeb-DF~\cite{li2020celebdflargescalechallengingdataset}) and third generation datasets (e.g., DFDC~\cite{dolhansky2020deepfakedetectionchallengedfdc}) further enhance the quantity, authenticity, and diversity of data. The latest third-generation datasets primarily focus on face-swapping methods (e.g., FSGAN, Faceswap), leading to homogenized artifacts, audiovisual mismatches, and poor generalization to emerging threats.

Crucially, as diffusion models revolutionize synthetic media~\cite{chen2024echomimiclifelikeaudiodrivenportrait,xu2024hallohierarchicalaudiodrivenvisual, cui2024hallo2longdurationhighresolutionaudiodriven, wang2024vexpressconditionaldropoutprogressive}, existing datasets fail to capture the multimodal realism and temporal coherence of modern digital human forgeries. This has caused detection systems to approach performance saturation on current benchmarks while remaining fragile in real-world scenarios. There is an urgent need for more challenging and representative datasets and benchmarks to address emerging threats.

To bridge this gap, we introduce \textbf{DigiFakeAV}, the new large-scale multimodal benchmark for detecting diffusion-based digital human forgeries. Our dataset advances the field through three key innovations:


$\bullet$  \textbf{Diffusion-Driven Synthesis:} unlike previous benchmarks relying on GANs or face-swapping, DigiFakeAV employs five state-of-the-art diffusion models (Sonic~\cite{ji2024sonicshiftingfocusglobal}, Hallo~\cite{xu2024hallohierarchicalaudiodrivenvisual}, etc.) to generate 60,000 high-resolution videos (8.4 million frames). These models leverage multimodal signals to create forgeries, bypassing source-target similarity requirements while achieving photorealistic details.

$\bullet$  \textbf{Multimodal Consistency:} we ensure strict alignment between synthesized lip movements, facial expressions, and head poses with audio prosody and semantic content through speech-driven audiovisual synchronization. Lip synchronization is significantly improved compared with the previous datasets based on Wav2Lip.

$\bullet$  \textbf{Scene-Aware Diversity:}  DigiFakeAV covers various real-world scenarios (news broadcasts, social media vlogs, multilingual interviews) with balanced demographic representation. We also increase the proportion of Asians and Africans, reducing bias while enhancing the robustness of cross-cultural detection.

Experiments reveal significant vulnerabilities in current detectors on DigiFakeAV, with state-of-the-art models showing marked AUC drops (e.g., SSVF~\cite{feng2023selfsupervisedvideoforensicsaudiovisual} shows a 43.5\% reduction). These results underscore the urgent need for robust detectors against next-generation forgeries. To address the challenge of diffusion-based digital human forgeries, we propose DigiShield, which is a new multimodal framework. By leveraging cross-modal attention and spatiotemporal feature extraction, it effectively identifies subtle inconsistencies between video and audio. Experiments demonstrate DigiShield’s superior performance on the DigiFakeAV dataset, highlighting its effectiveness for new deepfake detection and offering scalable insights for future research.

In summary, we conclude our work as follows:

$\bullet$ We introduce DigiFakeAV, a new large-scale multimodal benchmark for detecting diffusion-based digital human forgeries, comprising 60,000 high-resolution videos (8.4 million frames) generated by state-of-the-art diffusion models.

$\bullet$ Comprehensive experiments demonstrate that detectors suffer a large performance drop on DigiFakeAV,  highlighting the urgency of advanced detection methods.

$\bullet$ We propose DigiShield, a spatiotemporal multimodal framework that identifies deepfakes via cross-modal inconsistency learning, establishing a new baseline for detecting next-level deepfakes.

\begin{table}[tbp]
  \centering
  \caption{Quantitative comparison of the DigiFakeAV dataset with previous datasets}
  \label{tab:quantitativecomparison}
  \begin{adjustbox}{width=\columnwidth, totalheight=\textheight, keepaspectratio}
    \begin{tabular}{lccccccc}
      \toprule
      \textbf{Dataset}

        & \textbf{\shortstack{Real / Fake}}
        & \textbf{Total}
        & \textbf{Modality}
        & \textbf{Methods}
        & \textbf{Real Audio}
        & \textbf{Deepfake Audio} \\
      \midrule
      UADFV~\cite{yang2018exposingdeepfakesusing} (2018) & 49 / 49       &   98   & V   & 1 & No  & No  \\
      DF-TIMIT~\cite{korshunov2018deepfakesnewthreatface} (2018) & 320 / 640     &  960   & V   & 2 & Yes & No  \\
      FF++~\cite{rössler2019faceforensicslearningdetectmanipulated}(2019)  & 1,000 / 4,000 & 5,000  & V   & 4 & No  & No  \\

Google DFD~\cite{rössler2019faceforensicslearningdetectmanipulated} (2019) &   363 / 3,000 & 3,363  & V   & 5 & No  & No  \\

      Celeb-DF~\cite{li2020celebdflargescalechallengingdataset} (2020) &   590 / 5,639 & 6,229  & V   & 1 & No  & No  \\

DeeperForensics~\cite{jiang2020deeperforensics10largescaledatasetrealworld} ( 2020) &50,000 / 10,000& 60,000 & V   & 1 & No  & No  \\
      DFDC~\cite{dolhansky2020deepfakedetectionchallengedfdc} (2020) &23,654 /104,500&128,154 & A/V & 8 & Yes & Yes \\
      KoDF~\cite{Kwon_2021_ICCV} (2021) &62,166 /175,776&237,942 & V   & 6 & Yes & No  \\
      FakeAVCeleb~\cite{khalid2022fakeavcelebnovelaudiovideomultimodal} (2021) &   500 /19,500 &20,000  & A/V & 4 & Yes & Yes \\
      DeepSpeak~\cite{barrington2025deepspeakdataset} (2024) &6,226 / 6,799 &13,025  & A/V & 4 & Yes & Yes \\
      \midrule
      \textbf{DigiFakeAV (ours)} (2025) &10,000 /50,000&60,000  & A/V & 6 & Yes & Yes \\
      \bottomrule
    \end{tabular}
  \end{adjustbox}
\vspace{-6pt}

\end{table}

\section{DigiFakeAV: A Next Generation Deepfake Benchmark}

\begin{figure*}[!tbp]
    \centering
    \scalebox{0.8}{
        \includegraphics[width=0.83
        \linewidth]{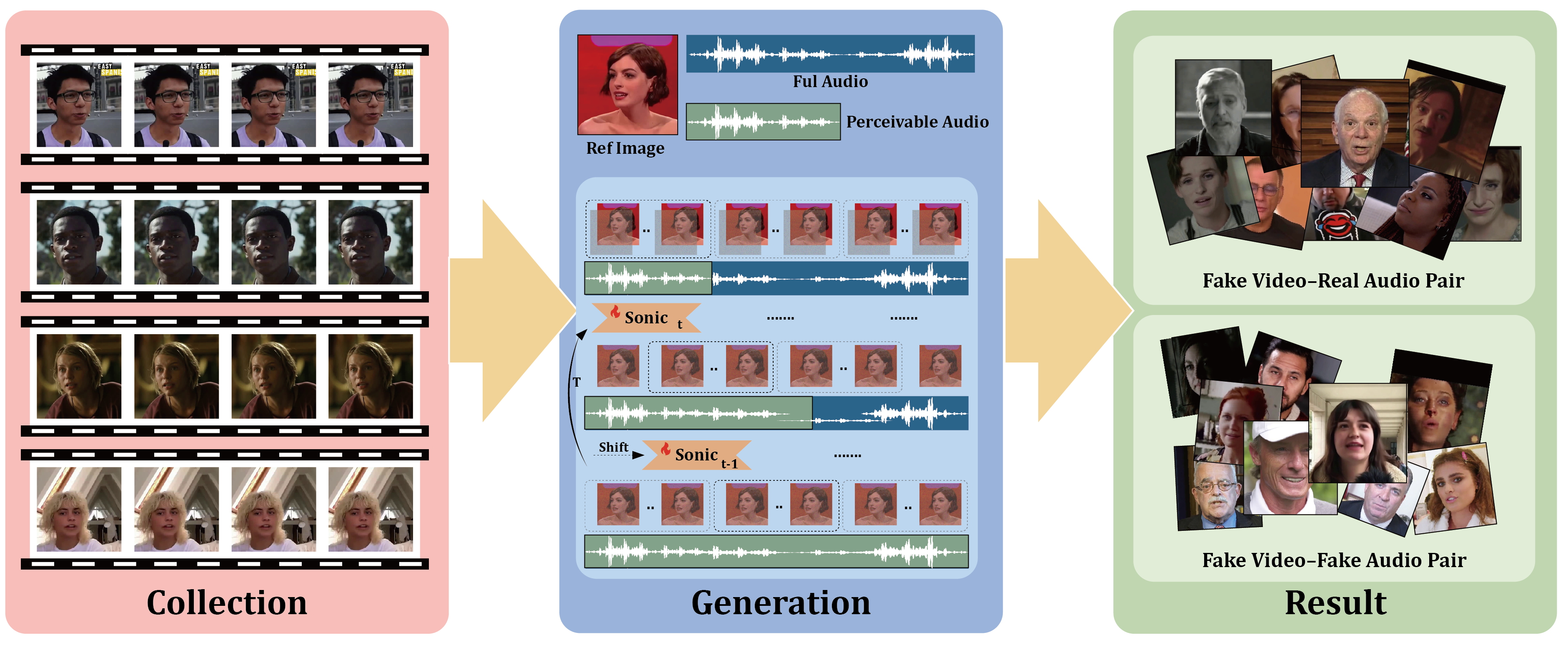}%
    }
    \vspace{-6pt}
    \caption{Workflow diagram for the construction of DigiFakeAV.}
    \label{fig:Workflow}

\vspace{-6pt}

\end{figure*}

In this section, we describe the pipeline of dataset construction, including data collection, data synthesis, and the composition of the final dataset, as shown in Fig. \ref{fig:Workflow}. 

\vspace{-10pt}
\subsection{Data Collection and Synthesis}

We curate original videos from the HDTF and CelebV-HQ datasets, applying standard pre-processing such as cutting, cropping, and re-encoding. Then, we produce the DigiFakeAV dataset using state-of-the-art video and audio synthesis models. Specifically, we employ five recent diffusion-based digital human video generation methods (Sonic~\cite{ji2024sonicshiftingfocusglobal}, Hallo~\cite{xu2024hallohierarchicalaudiodrivenvisual}, Hallo2~\cite{cui2024hallo2longdurationhighresolutionaudiodriven}, EchoMimic~\cite{chen2024echomimiclifelikeaudiodrivenportrait}, and V-Express~\cite{wang2024vexpressconditionaldropoutprogressive}) and one advanced audio synthesis method (CosyVoice 2~\cite{du2024cosyvoice2scalablestreaming}). As summarized in Tab.~\ref{tab:quantitativecomparison}, compared to previous deepfake methods, DigiFakeAV achieves third-generation scale, supports multimodal (video and audio) forgeries, and incorporates six new synthesis techniques.

\textbf{Synthesis Pipeline.} Both methods generate fake video and fake audio samples through three stages: (1) \textbf{Generation Conditioning:} the video generation model is driven by real audio (from real video–real audio clips) or synthetic audio (e.g., TTS), paired with static portraits or video frames as visual inputs. (2) \textbf{Diffusion Sampling:} video frames are synthesized via iterative denoising guided by audio and facial priors. Clip lengths range from 5 to 10 seconds, meeting the dataset’s resolution (512×512) and bitrate (3000 kbps) requirements. (3) \textbf{Quality Control:} samples are filtered using perceptual metrics (e.g., FID, Sync-C) to eliminate artifacts such as lip-sync errors or identity drift. Realistic noise and compression are added to simulate real-world distribution shifts.

\textbf{Key Advantages.}
(1) \textbf{Higher Realism:} diffusion models can generate fine-grained facial details (skin textures, eye movements) and maintain long-term temporal consistency, outperforming GAN-based methods.
(2) \textbf{Audiovisual Consistency:} Sonic and Hallo achieve strong lip-sync accuracy and emotional consistency, challenging unimodal detection.
(3) \textbf{Scalability:} both frameworks support batch processing and domain-specific fine-tuning (e.g., celebrity voice dubbing), enabling rapid dataset expansion.

By integrating these methods, DigiFakeAV provides rich fake video and fake audio samples that closely resemble real-world forgery scenarios, thus supporting the development of robust deepfake detection systems.

\vspace{-10pt}

\subsection{Dataset Organization}

Forged videos and audio can form four combinations: (1) Real Video–Real Audio (RV-RA), (2) Real Video–Fake Audio (RV-FA), (3) Fake Video–Real Audio (FV-RA), and (4) Fake Video–Fake Audio (FV-FA). We exclude RV-FA for two reasons: first, we focus on video-based facial forgery detection, whereas RV-FA does not involve human facial manipulation. For example, DFDC~\cite{dolhansky2020deepfakedetectionchallengedfdc} labels RV-FA samples as fake, which is inherently unfair to vision-only detectors. Second, audio-only manipulations without visual changes are often obvious and pose limited challenges for detection. For instance, FakeAVCeleb~\cite{khalid2022fakeavcelebnovelaudiovideomultimodal} has audio-video synchronization delays, which are relatively easy to detect. We then introduce three combinations and the corresponding methods adopted, with detailed explanations:

\textbf{Real Video–Real Audio (RV-RA).} From nearly 40,000 clean video clips, we select 10,000 real videos representing diverse ethnicities, genders, and ages. These videos are further processed to crop portrait regions, forming the basis of our real video dataset.

\textbf{Fake Video–Real Audio (FV-RA)} This category contains fake videos generated by real audio-driven diffusion models. We first extract and convert audio from real videos into the WAV format. RetinaFace is then used to detect and crop representative facial frames. Using five digital human generation methods: Sonic~\cite{ji2024sonicshiftingfocusglobal}, Hallo1~\cite{xu2024hallohierarchicalaudiodrivenvisual}, Hallo2~\cite{cui2024hallo2longdurationhighresolutionaudiodriven}, Echomimic~\cite{chen2024echomimiclifelikeaudiodrivenportrait}, and V-Express~\cite{wang2024vexpressconditionaldropoutprogressive}, we generate 25,000 forged videos. 

\textbf{Fake Video–Fake Audio (FV-FA).} This category encompasses both forged audio and video. We employ a pioneering voice cloning method, CosyVoice 2~\cite{du2024cosyvoice2scalablestreaming}. Specifically, we first use a large language model to generate processed text. By providing real audio-text pairs, the CosyVoice 2 model synthesizes fake audio mimicking the target speaker’s voice characteristics. We then generate 25,000 forged videos based on this fake audio using algorithms like Sonic, Hallo1, Hallo2, and Echomimic. 

\vspace{-10pt}
\subsection{Dataset Statistics}

Previous datasets such as HDTF~\cite{zhang2021flow} and CelebV-HQ~\cite{zhu2022celebvhqlargescalevideofacial} exhibit demographic imbalances in data distributions. For example, HDTF contains a male-to-female ratio of 65:35, while CelebV-HQ includes predominantly Caucasian subjects. To address this, we prioritize diversity across four dimensions during video selection: (1) \textbf{Gender Balance:} we achieve a 57:43 male-to-female ratio to reflect real-world demographics and avoid excessive gender skew. (2) \textbf{Ethnic Diversity:} the dataset includes White, African American, and Asian individuals, ensuring broad racial representation and enhancing model generalization. (3) \textbf{Nationality and Cultural Background:} subjects from diverse countries and cultural contexts are selected to reduce regional bias and enhance global representativeness. (4) \textbf{Scenario Diversity:} video content simulates real-world scenarios, including social interactions, news broadcasts, and daily conversations, to improve applicability in practical scenarios.

\vspace{-10pt}

\section{DigiShield: A Multimodel Deepfake Detection Baseline}

Existing models exhibit significant performance drops on the DigiFakeAV dataset (see Sec.~\ref{baselinecom}), underscoring limitations in multimodal detection of diffusion-based digital human forgeries. To address this, we propose DigiShield, which provides a new baseline for future research.

\subsection{Model Framework}
\begin{figure}[tbp]
  \centering
  \includegraphics[width=0.9\linewidth]{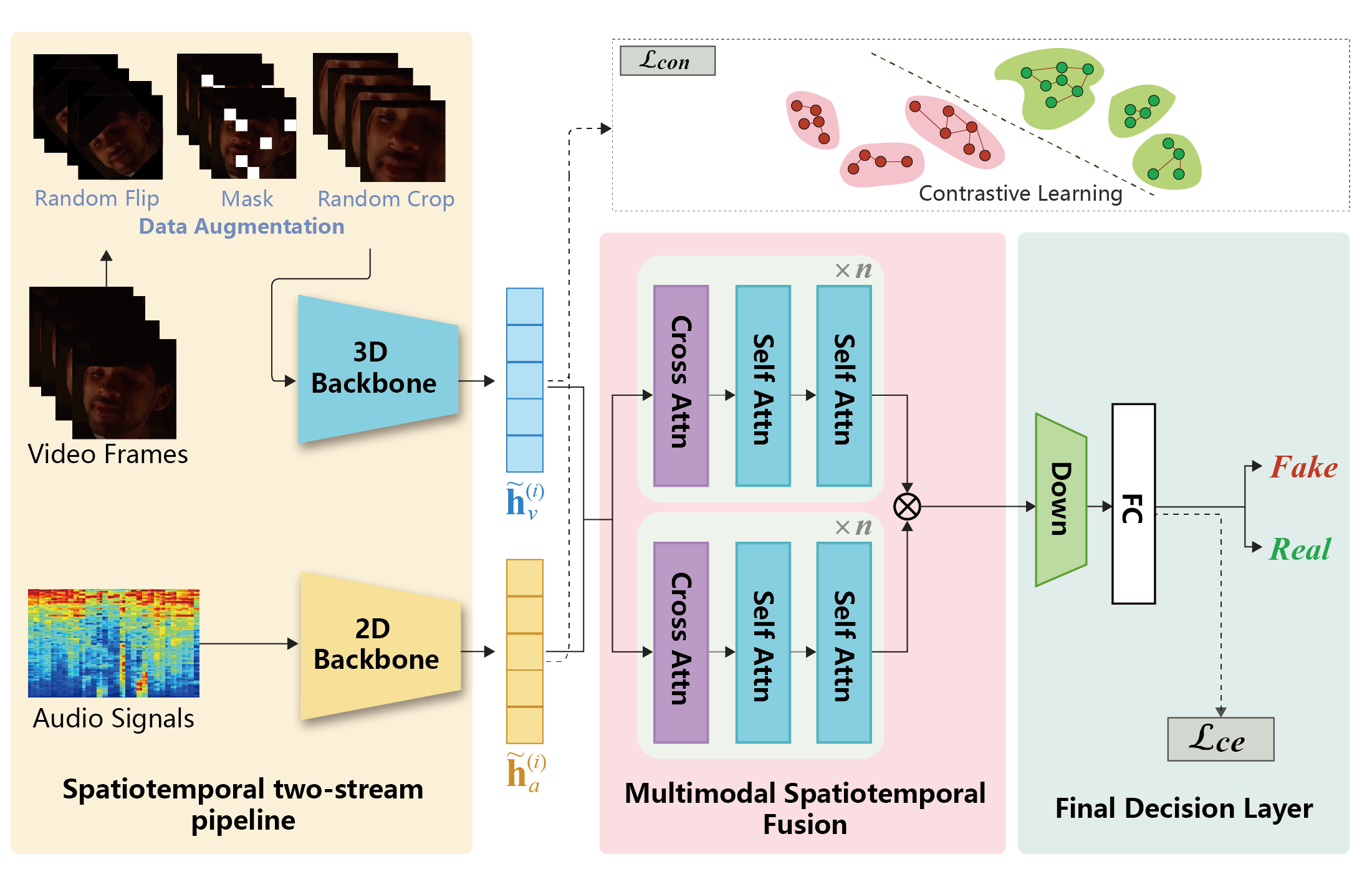}
 \vspace{-8pt}
  \caption{Overview of the DigiShield, it detects forgery traces by integrating spatiotemporal information from both video and audio modalities.}
  \label{DigiShield model}
  \vspace{-10pt}
\end{figure}

    We incorporate multimodal audio-visual cues. As shown in Fig.~\ref{DigiShield model}, our method includes (1) a spatiotemporal two-stream pipeline, (2) audio-visual fusion, and (3) a classification layer.

\textbf{Spatiotemporal Two-Stream Pipeline.} Formally, given video clips $x_{\mathrm{in}}^{v} \in \mathbb{R}^{T \times C \times H_v \times W_v}$ and audio signals $x_{\mathrm{in}}^{a} \in \mathbb{R}^{C \times H_a \times W_a}$, where $T$ is frame count, $C$ is the shared channel dimension, $H_v \times W_v$ video resolution, and $H_a \times W_a$ the Mel-spectrogram’s frequency and time bins, we use two-stream pipeline to extract:
$
f_v = \mathcal{F}_v\big(x_{\mathrm{in}}^{v}\big) \in \mathbb{R}^{T' \times C' \times H'_v \times W'_v}, \quad
f_a = \mathcal{F}_a\big(x_{\mathrm{in}}^{a}\big) \in \mathbb{R}^{C' \times H'_a \times W'_a}
$
where $\mathcal{F}_v$, $\mathcal{F}_a$ are visual/audio encoders, $T'$, $C'$, $H'_v \times W'_v$, and $H'_a \times W'_a$ are output temporal, shared channel, spatial, and frequency-time dimensions after convolution/pooling.

\textbf{Multimodal Spatiotemporal Fusion.} 
The classification layers are redesigned to extract spatiotemporal features for cross-modal fusion. Specifically, the extracted feature maps from both the video and audio streams, denoted as $f_v$ and $f_a$,  are reshaped into flattened representations:
$
\widetilde{f}_v \in \mathbb{R}^{N_v \times d} $ and
$\widetilde{f}_a \in \mathbb{R}^{N_a \times d},
$
where $ N_v = T' \times H'_v \times W'_v $ and $ N_a = H'_a \times W'_a $, with $ d = C' $ indicating the dimension of the feature channel.We then employ cross-attention and self-attention modules to enhance multimodal interaction and feature fusion as follows:
$\mathbf{z}_{va} = \operatorname{CrossAtt}(\mathbf{Q}=\widetilde{f}_v,\,\mathbf{K}=\widetilde{f}_a,\,\mathbf{V}=\widetilde{f}_a),\ 
 \mathbf{z} = \operatorname{SelfAtt}(\mathbf{Q}=\mathbf{z}_{va},\,\mathbf{K}=\mathbf{z}_{va},\,\mathbf{V}=\mathbf{z}_{va}).$

\textbf{Final Decision Layer.} Audio and video features are concatenated, and the resulting representation is fed into a fully connected layer for binary classification.

\vspace{-8pt}

\subsection{Loss Function}
The model is trained using both contrastive and cross-entropy losses to capture audio-visual inconsistencies indicative of forgery. (1) We adopt a contrastive loss defined as
\(\mathcal{L}_{\text{con}} = \tfrac{1}{N} \sum_{i=1}^{N} \big[ y_i D(\widetilde{\mathbf{h}}_v^{(i)}, \widetilde{\mathbf{h}}_a^{(i)})^2 + (1-y_i) \max(0, m - D(\widetilde{\mathbf{h}}_v^{(i)}, \widetilde{\mathbf{h}}_a^{(i)}))^2 \big]\),
where \(N\) is the batch size, \(\widetilde{\mathbf{h}}_v^{(i)}\) and \(\widetilde{\mathbf{h}}_a^{(i)}\) are the flattened visual and audio features of the \(i\)-th sample, \(y_i \in \{0,1\}\) denotes ground-truth labels, \(D(\cdot,\cdot)\) is the Euclidean distance (e.g., \(D(\mathbf{x},\mathbf{y})=\|\mathbf{x}-\mathbf{y}\|_2\)), and \(m>0\) is a margin parameter. The contrastive loss enforces compactness of audio-visual embeddings for genuine pairs while ensuring a margin-based separation for mismatched pairs. (2) The classification is guided by cross-entropy loss
\(\mathcal{L}_{\text{ce}} = - \tfrac{1}{N} \sum_{i=1}^N \big[ y_i \log \hat{y}_i + (1-y_i)\log(1-\hat{y}_i) \big]\),
where \(\hat{y}_i \in [0,1]\) is the predicted probability of the \(i\)-th sample being genuine. The overall loss is then
\(\mathcal{L}_{\text{total}} = \mathcal{L}_{\text{con}} + \mathcal{L}_{\text{ce}}\).

\vspace{-6pt}

\section{Experiment}
\label{exp}
\subsection{Implementation Details} 
During preprocessing, videos are processed with RetinaFace for facial detection and cropping, while audio tracks are resampled to 16 kHz. For DigiShield, we use ResNet-50 as the backbone. On DigiFakeAV, experiments adopt an 8:1:1 train/val/test split with no identity overlap, sampling 30 frames per video and applying data augmentation like random cropping. The Area Under the Curve (AUC) is the primary evaluation metric, with all preprocessing and training settings kept consistent for fair comparison. We also use DF-TIMIT~\cite{korshunov2018deepfakesnewthreatface} and process the dataset following the official settings.

\begin{table}[!t]
    \centering
    \caption{AUC scores (\%) of various methods on previous datasets and our DigiFakeAV.}
    \label{table:comparisonresults}
    \begin{adjustbox}{width=\linewidth, center}
    \begin{tabular}{@{}lcccccccc@{}}
        \toprule
        \multirow{2}{*}{\textbf{Method}} &  
        \multicolumn{2}{c}{\textbf{DF-TIMIT}} & 
        \multirow{2}{*}{\textbf{FF-DF}} &  
        \multirow{2}{*}{\textbf{DFDC}} & 
        \multirow{2}{*}{\textbf{Celeb-DF}} & 
        \multirow{2}{*}{\textbf{FakeAVCeleb}} & 
        \multirow{2}{*}{\textbf{DigiFakeAV}} \\
        \cmidrule(lr){2-3}
          & \textbf{LQ} & \textbf{HQ} & & & & &\textbf{(ours)} & \\
        \midrule
        Meso4   ~\cite{Afchar_2018}         & 87.8 & 68.4 & 84.7 & 75.3 & 54.8 & 60.9 & 50.1 \\
        MesoInception4 ~\cite{Afchar_2018}  & 80.4 & 62.7 & 83.0 & 73.2 & 53.6 & 61.7 & 63.8 \\
        Xception-c23  ~\cite{rössler2019faceforensicslearningdetectmanipulated}   & 95.9 & 94.4 & 99.7 & 72.2 & 65.3 & 72.5 & 66.1 \\
        Capsule  ~\cite{nguyen2019usecapsulenetworkdetect}        & 78.4 & 74.4 & 96.6 & 53.3 & 57.5 & 70.9 & 65.3 \\
        HeadPose   ~\cite{yang2018exposingdeepfakesusing}      & 55.1 & 53.2 & 47.3 & 55.9 & 54.6 & 49.0 & 51.7 \\
        F3-Net~\cite{qian2020thinkingfrequencyfaceforgery}           & 99.8 & 99.4 & 93.7 & 95.1 & 86.7 & 91.3 & 66.4 \\
        Cross Efficient ViT~\cite{Coccomini_2022} & 50.4 & 55.8 & 99.1 & 95.1 & 86.7 & 80.5 & 52.2 \\
        SSVF ~\cite{feng2023selfsupervisedvideoforensicsaudiovisual}            & -      & -      & -     & -     & -     & 94.5 & 51.0 \\
        SFIConv    ~\cite{10286083}      & 100.0 & 100.0 & 95.9 & 96.7 & 95.8 & 93.0 & 71.2 \\
        \bottomrule
    \end{tabular}
    \end{adjustbox}
\vspace{-10pt}

\end{table}

\vspace{-10pt}

\subsection{Baseline Comparison} 
\label{baselinecom}
To conduct a thorough evaluation of the DigiFakeAV dataset, we systematically select nine representative deepfake detection algorithms based on the criteria of representativeness and accessibility. The selected models cover six categories: Naive (Meso4~\cite{Afchar_2018}, MesoInception4~\cite{Afchar_2018}, Xception-c23~\cite{rössler2019faceforensicslearningdetectmanipulated}), Spatial (Capsule~\cite{nguyen2019usecapsulenetworkdetect}, HeadPose~\cite{yang2018exposingdeepfakesusing}), Frequency (F3-Net~\cite{qian2020thinkingfrequencyfaceforgery}), Temporal (Cross Efficient ViT~\cite{Coccomini_2022}), Hybrid Domain (SFIConv~\cite{10286083}), and Multi-modal (SSVF~\cite{feng2023selfsupervisedvideoforensicsaudiovisual}).

As shown in Tab.~\ref{table:comparisonresults}, existing deepfake detectors collapse on DigiFakeAV (e.g., SFIConv AUC drops from 100.0\% (DF-TIMIT HQ) to 71.2\%, SSVF from 94.5\% (FakeAVCeleb) to 51.0\%). This underscores the challenge from diffusion-based forgeries and the inadequacy of traditional texture or static frequency methods, highlighting the need for spatiotemporal and cross-modal analysis.

\vspace{-6pt}
\subsection{DigiShield Comparison}

As shown in Tab.~\ref{tab:method_comparison}, DigiShield achieves an AUC of 80.1\% on the challenging DigiFakeAV dataset, outperforming the best hybrid
domain method SFIConv (71.2\%) by 8.9\% and the frequency baseline F3-Net (66.4\%) by 13.7\%. This performance demonstrates DigiShield’s superior ability to capture subtle diffusion-based forgery artifacts through explicit spatiotemporal and multimodal learning. On the traditional DF-TIMIT dataset, DigiShield maintains a perfect 100\% AUC, validating its strong generalization. These results show that modeling multimodal spatiotemporal inconsistency enables effective and robust detection of digital human forgeries.

\begin{table}[tbp]
\centering
\caption{Comparison of AUC scores for DigiShield on DigiFakeAV and DF-TIMIT}
\label{tab:method_comparison}

\scalebox{0.8}{
\begin{tabular}{lccc}
\toprule
Method & DigiFakeAV & DF-TIMIT-LQ & DF-TIMIT-HQ  \\
\midrule
MesoInception4 & 63.8 & 80.4 & 62.7  \\
Capsule    & 65.3   & 78.4 & 74.4  \\
Xception-c23   & 66.1  & 95.9 & 94.4  \\
F3-Net    & 66.4       & 99.8 & 99.4  \\
SFIConv    & 71.2  & 100.0 & 100.0  \\
\midrule
\textbf{DigiShield (ours)}   & \textbf{80.1} & \textbf{100.0} & \textbf{100.0}  \\
\bottomrule
\end{tabular}}
\vspace{-6pt}

\end{table}

\begin{table}[tbp]
\centering
\caption{Ablation study of DigiShield in DigiFakeAV.}
\small 
\begin{tabular}{cccc|c}
\toprule
$\mathcal{L}_{ce}$ & $\mathcal{L}_{con}$ & CrossAtt & SelfAtt & AUC \\
\midrule
\checkmark & & & &  73.6 \\
\checkmark & \checkmark & \checkmark & &  77.4  \\
\checkmark & \checkmark & \checkmark & \checkmark &  \textbf{80.1} \\
\bottomrule
\end{tabular}
\label{tab:ablation}
\vspace{-6pt}

\end{table}

\vspace{-6pt}

\subsection{Ablation Study}

As shown in Tab.~\ref{tab:ablation}, ablation study validates our audio-visual spatiotemporal fusion strategy:
(1) The single-stream baseline (only use visual stream, $L_{ce}$) yields AUC 73.6\%.
(2) Introducing the audio stream and contrastive loss ($L_{con}$) raises AUC to 77.4\%, with $\text{CrossAtt}$ addressing cross-modal discrepancies. (3) Incorporating $\text{SelfAtt}$ further improves temporal coherence, achieving AUC 80.1\%. 
These results emphasize the importance of both temporal and cross-modal alignment for robust forgery detection.

\vspace{-8pt}

\section{Conclusion}
DigiFakeAV is a new multimodal dataset based on advanced digital human forgery technologies. As a pioneering dataset emphasizing diffusion-based synthesis, it introduces unprecedented challenges for deepfake detection. We evaluate its difficulty using nine detection methods, showing significant performance degradation compared to previous benchmarks.
Additionally, DigiShield, an audio-visual fusion baseline, is proposed to attempt to address this challenge.
DigiFakeAV and DigiShield aim to enable robust deepfake detection and inspire new research directions.

\bibliographystyle{IEEEbib}
\bibliography{strings,refs}
\end{document}